\documentclass[letterpaper, 10 pt, conference]{ieeeconf}  

\IEEEoverridecommandlockouts                              
\IEEEoverridecommandlockouts                              

\usepackage{graphicx} 

\title{\LARGE \bf
Triplet2Track: A Hierarchical System with Object-Centric Representations for Reliable Long-Horizon Manipulation
}

\author{
Jianxiang Liu$^{1}$,
Gaojing Zhang$^{2}$,
Chuan Wen$^{1}$,
Qipeng Liu$^{1}$,
Yuxuan Zhao$^{1}$,
Ning Guo$^{1}$,
Wenzhao Lian$^{1,*}$%
\thanks{This work was supported by the School of Artificial Intelligence, Shanghai Jiao Tong University.}%
\thanks{$^{1}$School of Artificial Intelligence, Shanghai Jiao Tong University, Shanghai, China.}%
\thanks{$^{2}$School of Engineering and Informatics, University of Sussex, Brighton, U.K.}%
\thanks{$^{*}$Corresponding author: { lianwenzhao@sjtu.edu.cn}.}%
}

\usepackage{amsmath}    
\usepackage{amssymb}
\usepackage{cite}
\usepackage{float}      
\usepackage{multirow}   
\usepackage{graphicx,cuted} 
\usepackage{array}      
\usepackage{balance}
\usepackage{booktabs}   
\usepackage{booktabs,multirow,subcaption}
\usepackage{caption}
\usepackage{tabularx}
\usepackage{array}
\usepackage{booktabs}

\makeatletter

\newcommand{\Rmnum}[1]{\expandafter\@slowromancap\romannumeral #1@}
\makeatother

\begin{document}


\maketitle

\vspace*{-50mm}
\begingroup\setlength\stripsep{0pt} 
\vspace*{-55mm}
\begin{strip}
\centering

\includegraphics[page=1,trim=80mm 200mm 6mm 20mm,clip,width=\linewidth]{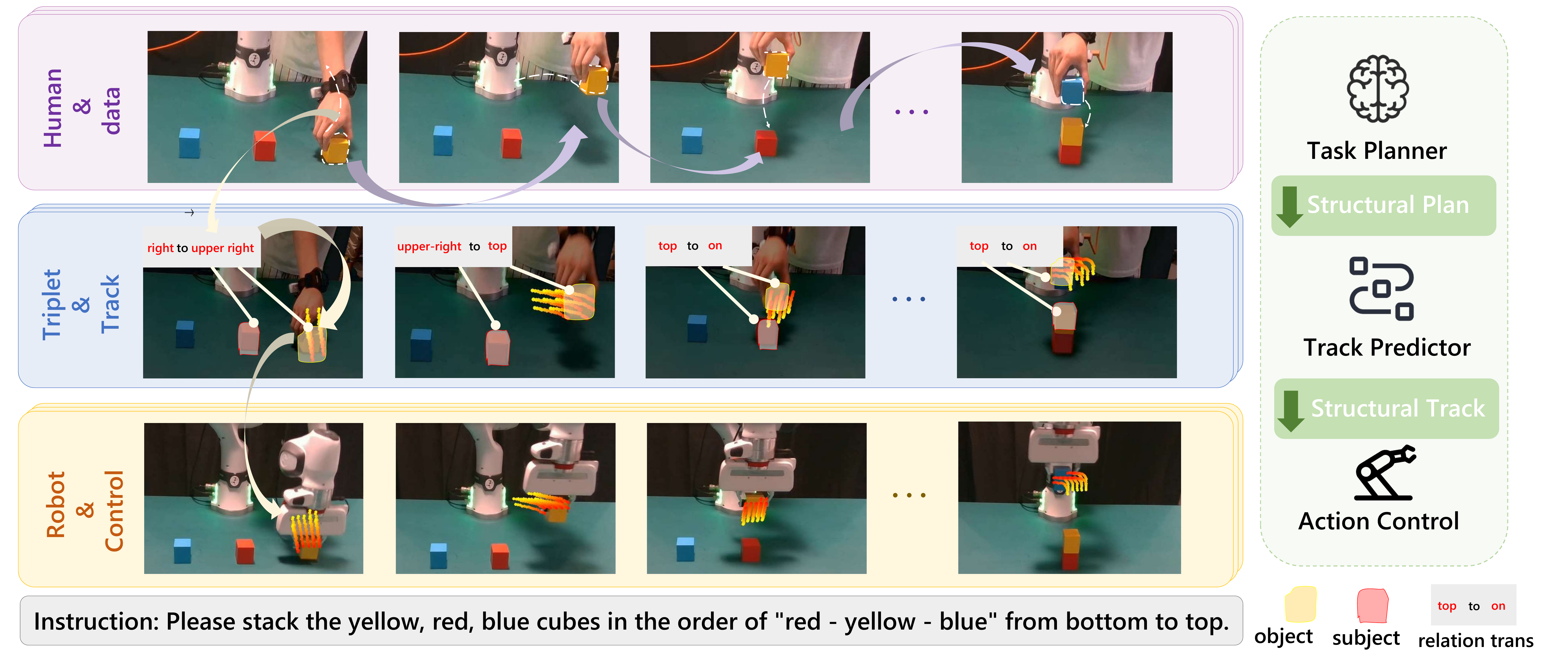}
\captionof{figure}{TTS is a hierarchical closed-loop system for reliable long-horizon manipulation, using instance-grounded triplets as structural task representations for planning and continuous point tracks as physical motion representations for execution. This design reduces planning hallucinations and supports data-efficient manipulation under limited robot data.}
\label{fig:two_columns}
\end{strip}
\endgroup

\vspace*{-3mm} 

\thispagestyle{empty}

\begin{abstract}
Ensuring reliability in uncertain environments remains difficult for long-horizon robotic manipulation. End-to-end VLA models are data-heavy and opaque, making diagnosis and verification difficult. Hierarchical pipelines are more interpretable, but their plans are often weakly grounded in observations, weakly aligned with low-level actions, and computed without online feedback, leading to open-loop behavior and hallucinations. To address these issues, we introduce the Triplet-to-Track System (TTS), a closed-loop long-horizon imitation learning system that uses human videos to reduce reliance on robot-collected data. TTS represents high-level subgoals as instance-grounded triplets, translates them into continuous track priors for execution, and monitors task progress from observations for online replanning. Across diverse real-world long-horizon tasks, TTS achieves a 74.8\% average success rate and supports object-level and compositional generalization.
\end{abstract}



\section{Introduction}

Long-horizon manipulation arises in many everyday tasks~\cite{liu2025aligning,duan2022survey,wang2024karma}, motivating the need for general policies that can reliably execute extended behaviors.
Compared with short-horizon settings with explicit goals and near-linear structure~\cite{yu2020meta,james2020rlbench,zeng2021transporter}, these tasks often involve goals that are only partially specified in the initial observations and become operationally verifiable only after a sequence of intermediate interactions~\cite{yang2025lohovlaunifiedvisionlanguageactionmodel}.
The central challenge is therefore not merely reactive control, but maintaining plans that remain correct and consistent with evolving observations over extended horizons.
A promising direction is to use goals that are directly checkable from visual observations and to monitor execution online so that deviations can be detected before errors accumulate~\cite{feng2025reflective,zhou2025step}.
Existing methods along this direction mainly fall into two paradigms: end-to-end visual--language--action (VLA) systems~\cite{black2024pi_0,zhang2024hirt,kim2024openvla} and hierarchical pipelines~\cite{wang2024vlm,myers2024policy,li2025gr}.

End-to-end visual--language--action (VLA) systems typically learn a direct mapping from raw images and language instructions to low-level actions, implicitly coupling perception, language understanding, planning, and control in a shared latent space. While this design avoids hand-crafted abstractions, it also forgoes structured, human-interpretable modeling of task semantics~\cite{black2024pi_0}. As a result, it has two main drawbacks. First, it exacerbates the long-standing data-efficiency challenge in imitation learning~\cite{wen2023any,brohan2022rt}. In long-horizon settings, extended episodes substantially increase supervision demands and make it far more costly to collect sufficiently diverse demonstrations than in short-horizon tasks~\cite{gao2025vla}. Second, because the reasoning process is internalized within the network, such policies are difficult to interpret and diagnose, which can lead to unreliable decisions at execution time~\cite{wang2025vlatest}.

Many hierarchical approaches reduce data requirements by combining human demonstrations with high-level planners built on pretrained LLMs or VLMs, thereby leveraging web-scale knowledge and reducing reliance on robot-collected data~\cite{wang2024vlm,myers2024policy,ahn2022can}. However, these approaches still have important limitations. QA-conditioned policies~\cite{wang2024vlm,ahn2022can} typically produce free-form language goals without explicit structural or spatial constraints, making hallucinated objects and relations common. Semantic decompositions~\cite{myers2024policy,wu2024mldt} introduce symbolic task structure, but remain weakly grounded in concrete object instances and scene states. As a result, when hallucinations or semantic drift occur, these systems often lack closed-loop mechanisms to verify subgoals against visual observations or revise inconsistent plans, leaving many failures undetected and unrecovered.


To jointly address the data inefficiency and limited verifiability of end-to-end VLAs, as well as the weak grounding and lack of automatic error detection and recovery in hierarchical schemes, we present the \textbf{T}riplet-to-\textbf{T}rack \textbf{S}ystem (\textsc{TTS}), a framework for reliable long-horizon imitation learning with limited robot data.
\textsc{TTS} uses discrete triplets for planning and continuous tracks for execution.
Specifically, it represents task goals with object-centric triplets that bind spatial relations to concrete object instances, and maps subgoal transitions to continuous tracks that low-level controllers can consume as physical priors~\cite{wen2023any}.
Unlike prior methods, TTS makes the planning–control interface explicit, instance- and spatially grounded, and visually verifiable.

    
    
Our contributions are three-fold:
\begin{enumerate}
    \item We introduce \textbf{T}riplet-to-\textbf{T}rack \textbf{S}ystem (\textsc{TTS}), a closed-loop long-horizon imitation learning system that uses human videos to reduce reliance on robot-collected data, while improving execution reliability through observation-based monitoring and replanning.
    
    \item We present a structured way to connect planning and control, where high-level subgoals are expressed as instance-grounded triplets and translated into continuous track priors for low-level execution, making task progress explicit and verifiable.
    
    \item We validate TTS on real-world long-horizon manipulation tasks under perturbation and limited-data settings, showing reliable performance together with generalization to unseen object instances and task compositions.
\end{enumerate}

The remainder of this paper reviews related work and problem formulation in Sections II–III, presents TTS in Section IV, reports real-world experiments in Section V, and concludes the paper in Section VI.

\section{Related work}

\subsection{Methods for Long-horizon tasks}

Long-horizon manipulation is commonly approached via two families of methods. End-to-end VLA systems map pixels and language directly to actions~\cite{black2024pi_0, kim2024openvla, cheang2024gr, zhang2024hirt, brohan2022rt, zhang2024vlabench, fan2025long}. These models can scale with data but remain opaque and data-hungry, and their behavior often degrades under distribution shift. Hierarchical pipelines instead use a high-level planner (typically a VLM) coupled with a low-level controller~\cite{wang2024vlm, myers2024policy, li2025gr, sun2024hierarchical, sundaresan2024s, zhou2024isr}. Planners usually emit semantic sketches, which improves interpretability but still risks semantic hallucination~\cite{ahn2022can, wang2024vlm} and suffers from planner--controller interface mismatches, especially in open-loop settings or under imperfect synchronization.

A complementary line decomposes tasks into atomic units in semantic or relational space~\cite{myers2024policy, wu2024mldt}. While these approaches increase data efficiency and modularity, their abstractions often remain semantic rather than instance-level and spatially grounded, limiting verifiability against concrete scene geometry and contact. Recent systems~\cite{yang2025lohovlaunifiedvisionlanguageactionmodel, zhou2025step} add periodic replanning or feedback, yet without predicates that are directly checkable in perception, closed-loop reliability can still be brittle. These limitations motivate a representation that is hierarchical, instance-grounded, visually verifiable, and directly connected to low-level execution.

\subsection{Policy learning from action-free videos}

Videos carry rich cues about behavior and scene dynamics~\cite{ shao2021concept2robot, shaw2023videodex}, but often lack action supervision. Prior work tackling this typically follows three routes: (i) inverse-dynamics pseudo labeling---data-efficient~\cite{baker2022video, torabi2018behavioral} but brittle for continuous actions and long horizons; (ii) self-supervised feature pretraining~\cite{nair2022r3m, ma2022vip, sermanet2018time}---useful for perception yet weak on temporal control structure; and (iii) video prediction or goal-image generation---promising~\cite{ko2023learning, bharadhwaj2023zero} but prone to hallucinated geometry or contact and high inference cost. An alternative is to predict \emph{point tracks} as physically grounded motion priors~\cite{wen2023any, bharadhwaj2024track2act, xu2024flow}, enabling closed-loop guidance with low compute. However, such track-only priors lack global task-level context, so execution can become myopic and get trapped in local limit cycles. This motivates our use of tracks not as a standalone execution prior, but as a task-conditioned interface driven by explicit and verifiable high-level subgoals.

\begin{figure*}[t] 
  \centering
  \includegraphics[
    page=1,
    trim=5mm 380mm 320mm 2mm,clip,
    width=\textwidth
  ]{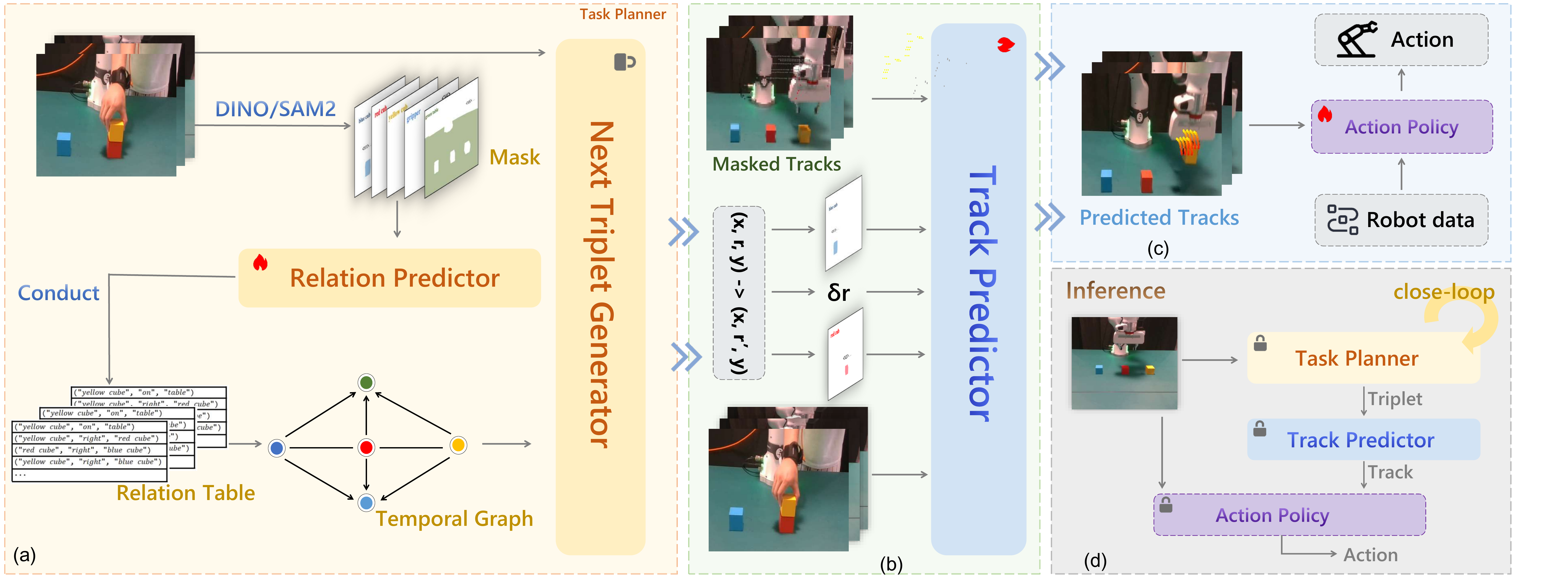} 
  \caption{: TTS pipeline. (a) The Relation Predictor extracts per-frame triplets and constructs a temporal graph, while the Next-Triplet Generator selects the next target triplet. (b) The Track Predictor converts the target transition into continuous point tracks. (c) The Action Policy maps tracks to actions. (d) During inference, TTS detects the current triplets, selects a feasible target, predicts track priors, and executes actions in a closed loop.}
  \label{fig:two_columns}
\end{figure*}

\section{Preliminaries}

We model real-world long-horizon tasks as a contextual POMDP
\(\mathcal{M}_\ell = (\mathcal{S}, \mathcal{A}, P, \Omega, \mathcal{O}, \rho_0)\),
where at time \(t\), the state \(s_t \in \mathcal{S}\) produces an observation \(o_t \in \Omega\) via \(\mathcal{O}(o_t \mid s_t)\), after which the agent selects an action \(a_t \in \mathcal{A}\) using \(\pi(\cdot \mid o_{1:t}, \ell)\), and the next state follows \(P(s_{t+1} \mid s_t, a_t)\) with \(s_0 \sim \rho_0\).

Guided by this formulation, we decouple high-level goal specification from low-level execution and use two complementary representations.
The triplet space \(\mathcal{C}=\mathcal{I}\times\mathcal{R}\times\mathcal{I}\) contains predicates \(c=(x,r,y)\) over object instances \(\mathcal{I}\) and relations \(\mathcal{R}\), where $R$ is a predefined vocabulary of spatial and gripper–object relations, such as on, over, and grasped. The function \(\mathrm{hold}(o_t,c)\in\{0,1\}\) indicates whether relation \(r\) holds for the ordered instance pair \((x,y)\) in the current scene. 
The track space \(\mathcal{T}\) comprises entity-aligned motion descriptors \(\tau=\{z_t,\dots,z_{t+L-1}\}\), with \(z_i\in\mathbb{R}^d\), that parameterize temporally extended control.

Given \((o_t,\ell,c_k)\), the Task Planner proposes the next subgoal \(c_{k+1}\), and the Track Predictor maps the discrete transition \(c_k \rightarrow c_{k+1}\) to a continuous track \(\tau_k\), which the low-level policy uses to generate action \(a_{t:t+L-1}\sim\pi_L(\cdot\mid o_t, c_k, \tau_k)\).
The detailed closed-loop execution rule, including guard evaluation and replanning, is described in Sec.~IV.

\section{Method}
\label{sec:method}

In this work, we aim to build a robot control system that is reliable, robust, and data-efficient. We propose the Triplet-to-Track System (TTS), which organizes control into three layers: \textbf{a Task Planner} that turns human videos into discrete, validatable, instance-spatial triplets; \textbf{a Track Predictor} that compiles each triplet into a continuous, entity-aligned track; \textbf{an Action Policy} that uses the predicted track as a physical prior to produce actions. Triplets bind object pairs and relations to concrete physical instances and are directly checkable from observations, which helps reduce planning hallucinations and supports spatial reasoning. However, acting directly on triplets leaves a gap between discrete subgoals and continuous control. TTS bridges this gap with a Track Predictor that generates fine-grained, structured, and instance-grounded tracks. The Track Predictor is pretrained on action-free videos, which improves data efficiency under limited robot data~\cite{wen2023any, bharadhwaj2024track2act, xu2024flow}. 
Conditioning tracks on triplet priors further sharpens relational transformation learning and reduces sensitivity to appearance variations, improving object-level and compositional generalization under similar grasp poses. Together with online detection and replanning, these components form a closed-loop system for long-horizon manipulation, as shown in Fig.~2.

\subsection{Task Planner for discrete, abstract triplets}

We seek a task planner that (i) infers triplets among objects from an admissible set to extract an object-relation graph from the current frame, and (ii) proposes the next discrete subgoal as a triplet. Operationally, the planner forms a two-stage perception-to-decision pipeline: a \textbf{Relation Predictor} and a \textbf{Graph-Guarded Next Triplet Generator}.

\textbf{Relation Predictor.}
Reliable detection of spatial relations is challenging, as a universal zero-shot detector would require large-scale relation annotations, diverse data, and heavy pretraining~\cite{wen2024can, chen2024spatialvlm, cheng2024spatialrgpt}. Since our goal is to reduce planning hallucination and improve reliability, rather than train such a detector, we adopt a low-overhead design with two complementary modules: a \textbf{L}ightweight \textbf{T}ransformer-based \textbf{P}redictor (LTP) and a LoRA-tuned VLM branch. The LTP is a pairwise relation classifier built with four Transformer layers and two linear layers, taking per-instance SAM2 decoder tokens as input and predicting relation logits for each ordered object pair.

Using Grounding-DINO~\cite{liu2024grounding} and SAM2~\cite{ravi2024sam} for instance detection, we feed the per-instance tokens from the SAM2 decoder directly into the LTP. The LTP outputs a tensor \(P_t \in \mathbb{R}^{n\times n\times |\mathcal{R}|}\), where \(n\) is the number of detected object instances in the current frame, \((x,y)\) denotes an ordered object pair among these \(n\) instances, and \(\mathcal{R}\) is a predefined relation vocabulary. For any object pair \((x,y)\), the predicted relation is obtained by applying a softmax over relation logits: \(\hat r_{xy}^{\mathrm{LTP}}(t)=\arg\max_{r\in\mathcal{R}} P_t(x,y,r)\).

The dominant error of the LTP is \emph{spurious flips} on relations that should remain stable over short horizons, where the prediction changes for one frame and then quickly reverts. Formally, on adjacent frames \((t,t+1)\), we define \(\mathrm{FP}\) when \(r_{xy}(t)=r_{xy}(t+1)\) but \(\hat r_{xy}^{\mathrm{LTP}}(t)\neq \hat r_{xy}^{\mathrm{LTP}}(t+1)\) (spurious flip), and \(\mathrm{FN}\) when \(r_{xy}(t)\neq r_{xy}(t+1)\) but \(\hat r_{xy}^{\mathrm{LTP}}(t)=\hat r_{xy}^{\mathrm{LTP}}(t+1)\) (missing change). With per-frame accuracy \(p\) and per-step change rate \(\rho\), a two-frame analysis yields \(\mathbb{E}[\mathrm{FP}] \ge (1-\rho)\,2p(1-p)\) and \(\mathbb{E}[\mathrm{FN}] \le \rho(1-p^2)\), hence spurious flips dominate whenever \(\rho < \tfrac{2p}{1+3p}\). In practice, we observe \(p\in(0.80,0.90)\) and \(\rho\approx 0.02\), so \(\rho \ll \tfrac{2p}{1+3p}\), which matches the transient few-frame flips observed in practice. A naive sliding-window average over per-frame probability vectors reduces but does not remove this jitter.

We therefore fine-tune InternVL2.5~\cite{chen2024expanding} as a high-precision verifier and invoke it only when a change is suspected. This event-triggered VLM improves reliability while avoiding the cost of per-frame VLM inference. When fine-tuning the VLM, we exploit the structured, object-anchored form of triplets to classify within a masked relation set rather than generate free-form text: given \((x,y)\), we condition the prompt on their instance masks and restrict the label space to feasible relations for that pair. For training the LTP and the VLM branch, we use only five sequences per task. Since relation changes are sparse, we annotate only change points to keep supervision cost low.

In summary, inference follows a cascade: a high-rate spatial predictor monitors guarded pairs and raises a flag only when the windowed argmax changes, indicating a suspected transition. Upon a flag, the fine-tuned VLM verifier computes \(q(r\mid o_t,x,y,\ell)\)  to confirm the change. This high-recall detector together with high-precision verification preserves responsiveness while sharply reducing false positives, especially for context triplets that should remain unchanged.

\begin{figure}[t]  
  \centering
  \setlength{\belowcaptionskip}{-0.3cm} 
  \includegraphics[
    page=1,
    trim=0mm 223mm 120mm 25mm,clip,
    width=\textwidth
  ]{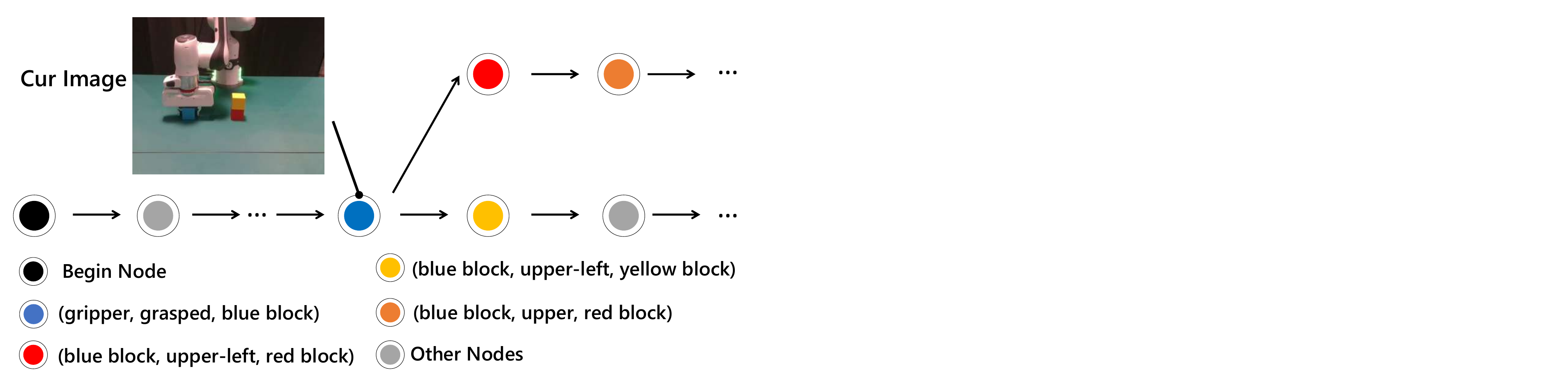} 
  \caption{An example of node transformation in a graph.}
  \label{fig:two_columns}
\end{figure}

\textbf{Graph-Guarded Next Triplet Generator.}
We use a zero-shot VLM (e.g., GPT\text{-}4o~\cite{islam2025gpt}) to propose the next subgoal. Unlike prior work, we constrain its proposals with a triplet-based temporal graph to reduce planning hallucinations. We build a directed graph \(G=(V,E)\), where each node is a task-relevant triplet \(c=(x,r,y)\) and each edge represents a feasible transition. The graph is constructed from action-free videos: at each frame \(t\), the Relation Predictor outputs a relation table \(R_t\); when two consecutive tables \(R_t\) and \(R_{t+1}\) differ by a transition \((x,r,y)\rightarrow(x,r',y)\), we add the edge \(c\rightarrow c'\) and merge duplicates across demonstrations. Each demonstration forms one execution path, while different execution orders add branches.

The resulting graph serves both as a progress index and as a feasibility filter. At step \(k\), suppose the current node is \(c_k=(x,r,y)\). Its outgoing edges enumerate the feasible one-step updates observed in human data, allowing only observed transitions and excluding invalid ones. We encode \(G\) by its adjacency matrix \(A\), and use the row \(A[i_k,:]\), where \(i_k=\mathrm{index}(c_k)\), as a binary mask \(\mathcal{M}_k\) over the candidate next nodes. The VLM receives \(\{o_t,\ell,c_k,\mathcal{M}_k\}\) through a single prompt template and scores only masked candidates; proposals outside the mask are ignored. In this way, open-ended generation is turned into verifiable instance-level selection, which keeps attention on grounded object relations and suppresses hallucinated transitions. The graph is extracted from human demonstrations rather than manually specified, and can be naturally expanded as additional human videos become available to cover new execution orders and transition patterns. For example, in Fig.~3, if \(c_k\) is the blue node, then the next step can choose only the red or yellow nodes, while all other nodes are excluded.

Inference starts from the initial node \(c_0\). During execution, the guard set maintains the latest validated relation for each ordered instance pair \((x,y)\) encountered in completed subgoals: when a subgoal \(c=(x,r,y)\) is completed, the entry \((x,y)\mapsto r\) is inserted if absent, and otherwise the previous relation for \((x,y)\) is replaced by \(r\). At each step, the Relation Predictor evaluates the current guard set and returns an indicator \(G_k\in\{0,1\}\), where \(G_k=1\) indicates that all guards hold. If \(G_k=1\), the next subgoal is proposed within the masked neighborhood of \(c_k\); otherwise, the planner relocalizes to \(c_0\) and replans.
Formally,
\[
c_{k+1}=
\begin{cases}
\mathrm{VLM}_{\mathrm{planner}}(o_t,\ell,c_k,\mathcal{M}_k), & G_k=1,\\[2pt]
c_0, & G_k=0.
\end{cases}
\]

\subsection{Track Predictor for continuous tracks}

To map discrete triplets to continuous tracks, we train a Track Predictor. Unlike track-only motion priors, our predictor is conditioned on triplet transitions, so the predicted tracks are tied to explicit object pairs and relation changes rather than inferred solely from local motion cues. Track representations also exhibit cross-embodiment capability~\cite{wen2023any, bharadhwaj2024track2act}, enabling pretraining on action-free videos and reducing the amount of action-labeled robot data required.

\textbf{Data preparation.}
For each sequence, we first derive the triplet transition and the aligned video segment, following the same procedure used to build the temporal graph in the Task Planner. Using Grounding-DINO~\cite{liu2024grounding} and SAM2~\cite{ravi2024sam}, we obtain pixel-level masks for \(x\) and \(y\), which are reused as inputs \((M_x,M_y)\) to explicitly localize the entities and suppress background and appearance distractions. Using the mask-derived boxes, we seed points inside each instance and track them with CoTracker~\cite{karaev2024cotracker3} to obtain entity-aligned trajectories \(V_{t:t+H}\). The relation change \(r\!\to\!r'\) is encoded as a discrete token via a text encoder, yielding training samples \(\mathcal{D}_{x:y}=\{(O_{t:t+H},M_x,M_y,V_{t:t+H},e(r\!\to\!r'))\}\). We refer to \((M_x,M_y,e(r\!\to\!r'))\) as the triplet-transition prior, which instantiates the discrete transition \((x,r\!\to\!r',y)\) for learning.

\textbf{Multimodal masked track prediction.}
Inspired by ATM~\cite{wen2023any}, we cast track forecasting as a multimodal masked prediction task. Given the visual observation \(o_t\), current point positions \(v_t\), instance masks \(M_x,M_y\), and a relation-delta embedding \(e(r\!\to\!r')\), the model predicts the next positions \(v_{t+1}\) for each query point. We encode \(M_x\) and \(M_y\) with a dedicated mask encoder to produce instance-anchored mask tokens, and embed \(e(r\!\to\!r')\) into the shared token space as a conditioning signal. Compared with formatting the same information as a generic text prompt, this triplet-conditioned design provides a stronger inductive bias and clearer spatial supervision for learning relation-dependent motion. All other preprocessing follows ATM~\cite{wen2023any}. The fused tokens are then processed by a Transformer backbone, and a trajectory decoder produces continuous, entity-aligned tracks that realize the discrete update \((x,r,y)\!\to\!(x,r',y)\) as a motion prior for the low-level policy.

To enhance stability, we introduce two auxiliary objectives. The first is a masked-patch reconstruction objective, in which a decoder reconstructs masked image patches from visual features processed by the Transformer backbone. The second is relation-change recognition: using the object and subject mask decoder features together with intermediate text decoder tokens, the model predicts the relation-change class \(\delta r\), strengthening its modeling of relation changes. Conditioning tracks on triplet-transition priors also helps reduce sensitivity to appearance variation and improve object-level and compositional generalization under similar grasp poses.

\subsection{Policy Learning for Action}

\begin{figure}  
  \centering
  \setlength{\belowcaptionskip}{-0.3cm}
  \includegraphics[
    page=1,
    trim=220mm 230mm 90mm 90mm,clip,
    width=\textwidth
  ]{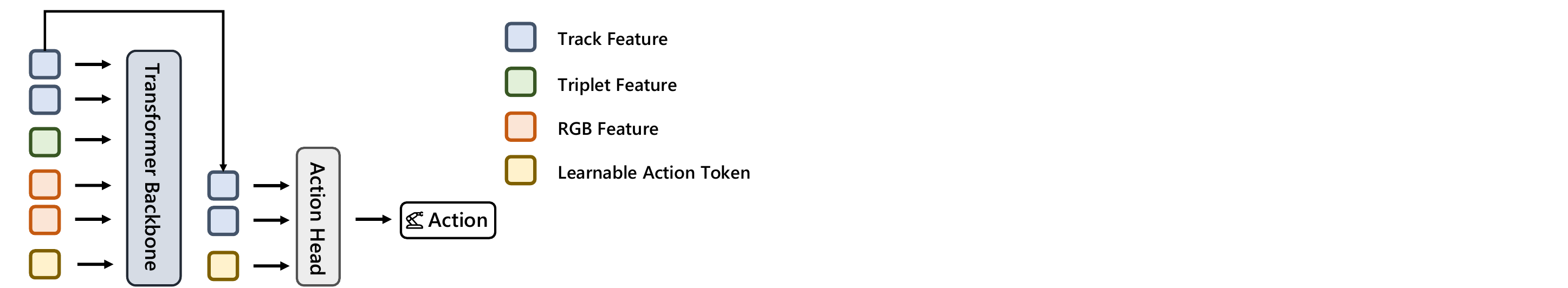} 
  \caption{The Action Policy consists of a Transformer backbone and an MLP-based Action Head.}
  \label{fig:two_columns}
\end{figure}

To map the predicted track prior to the robot action space, we train a track-guided behavior cloning policy on a small set of action-labeled robot data. Preprocessing follows the Track Predictor. We denote training samples by \(\mathcal{D}^{\mathrm{act}}_t=\{O_{t:t+H}, S_t, A_t, M_x, M_y, e\}\), where \(S_t\) denotes the joint and gripper states, and \(A_t\) is the action label consisting of a 7-dimensional end-effector pose and a scalar gripper width.

Given \((O_{t:t+H}, M_x, M_y, e)\), the Track Predictor first produces a short-horizon track prior \(p_{t:t+H}\). The low-level policy first fuses the observation features, triplet-transition prior, and track prior, and processes them using a Transformer backbone with a learnable action token. The resulting action token is then fused again with the track prior and decoded by an action head to predict the current robot action \(a_t\). In this way, the policy conditions action generation on both the local motion prior and the high-level relational context. The policy architecture is illustrated in Fig.~4. We train the policy with an \(L_2\) action-regression loss against \(A_t\).

Inference follows a receding-horizon scheme. At time \(t\), given \((o_t,\ell,c_k)\), the Task Planner sets \(c_{k+1}=(x,r',y)\) as the active goal. Every 5 steps, the Relation Predictor checks whether \(c_{k+1}\) has been achieved. If not, execution continues under the current goal; otherwise, the Next-Triplet Generator issues \(c_{k+2}\), which becomes the new active goal. Conditioned on the transition \(c_k\!\to\!c_{k+1}\), we compute a short-horizon track prior \(p_{t:t+H}\) from the two camera views, and the low-level policy consumes this prior together with the current observation to predict the current action \(a_t\). The action is executed, the observation is updated, and the horizon recedes. If a guard violation is detected, the planner relocalizes to the initial node of the graph and replans before continuing.

\section{Experiments}
We evaluate our approach through the following questions:

\textbf{Q1: Performance \& Generalization.}
Compared with representative baselines, does our model achieve strong performance on real-world long-horizon manipulation while maintaining high success rates under object-level and compositional generalization?

\textbf{Q2: Reliability \& Stability.}
Does TTS enable closed-loop, disturbance-robust execution through online replanning from observation feedback, while reducing hallucination and keeping goals and execution traceable through its structured, instance-spatial representation?

\textbf{Q3: Design Justification.}
Are the components of the Triplet-to-Track architecture necessary and well motivated, and what are their individual contributions?

\subsection{Experimental Setup}

\textbf{Tasks and datasets.}
We evaluate language-conditioned long-horizon manipulation on a physical 7-DoF Franka robot in the real world using three task families, each probing a different aspect of the system.

\begin{itemize}
    \item \textbf{Dynamic scale balancing (DSB).}
    This task evaluates closed-loop reasoning and online replanning in a two-pan balancing problem. A pile of toy frogs is placed in the left pan, and the goal is to balance the scale by grasping and placing the brown, blue, and red weights in the right pan. The task is always solvable. For training, we do not enumerate all balancing configurations. Instead, we collect videos containing primitive pick-and-place segments for each weight, so that the policy learns the track-to-action mapping while multi-step decision making is deferred to test-time planning.

    \item \textbf{Sequential stacking (SS).}
    This task evaluates whether the system can execute a prescribed multi-step order under rigid-body constraints. The goal is to stack the blocks from bottom to top in a specified order, such as red-yellow-blue. During training, the robot sees only one bottom-to-top order, while test-time evaluation requires generalization to unseen orderings.

    \item \textbf{Diverse pick-and-place (DPP).}
    This task evaluates goal understanding and object-level generalization. We consider two specifications: (i) placing the bitter melon into the blue box and the yellow potato into the green box, and (ii) placing the yellow, red, and pink cubes into the green box.
\end{itemize}

We label a task instance as \emph{Seen} if both its object set and execution order appear in the robot training data; otherwise, it is labeled \emph{Unseen}. The unseen setting is further divided into \emph{compositional} and \emph{cross-category} generalization. For \textbf{SS}, unseen evaluation requires generalizing from the training order ``red--yellow--blue'' to a novel order such as ``red--blue--yellow,'' which tests compositional generalization over the same object set. For \textbf{DPP}, unseen evaluation substitutes object categories or attributes at test time, e.g., replacing \emph{bitter melon} with \emph{corn} or \emph{red tomato}, which tests cross-category generalization. For \textbf{DSB}, after each pick-and-place step, if the scale remains unbalanced, the system replans the next move from the latest observation, which evaluates closed-loop planning under evolving task states.

Objects are randomly positioned within a designated workspace. The evaluated episodes span multi-step horizons, with DPP, SS, and DSB involving up to nine, eleven, and twelve subgoals, respectively. For each task, we collect 30 robot demonstrations and 50 human demonstrations, with 5 trajectory annotations for the robot and human videos, respectively. The human demonstrations contain the object variations and execution orders used for generalization, whereas the robot demonstrations do not.

\begin{table}[H]
  \centering
  \caption{Success rates on long-horizon tasks (\%).}
  \label{tab:merged_success_rates}
  \setlength{\tabcolsep}{3pt}
  \renewcommand{\arraystretch}{1.05}
  \resizebox{\columnwidth}{!}{%
  \begin{tabular}{@{}l cc cc ccc@{}}
    \toprule
    \multirow{2}{*}{Method} &
    \multicolumn{2}{c}{Dynamic Sacl-B} &
    \multicolumn{2}{c}{Seq-Stacking} &
    \multicolumn{3}{c}{Diverse Pick\&Place} \\
    \cmidrule(lr){2-3}\cmidrule(lr){4-5}\cmidrule(lr){6-8}
      & Seen & Perturbed
      & Seen & Unseen
      & Seen & Unseen & Perturbed \\
    \midrule
    FT-Pi0.5~\cite{intelligence2025pi_} & 13.3 & 0.0 & 25.0 & 0.0 & 40.0 & 16.7 & 40.0 \\
    SeeDo~\cite{wang2024vlm}            & 26.7 & 0.0 & 41.7 & 40.0 & 50.0 & 54.5 & 0.0 \\
    PALO~\cite{myers2024policy}         & 14.3 & 0.0 & 16.7 & 18.2 & 42.9 & 36.4 & 0.0 \\
    \textbf{TTS}                       & \textbf{78.6} & \textbf{60.0} &
                                         \textbf{83.3} & \textbf{66.7} &
                                         \textbf{84.6} & \textbf{70.0} & \textbf{80.0} \\
    \bottomrule
  \end{tabular}}
\end{table}


\textbf{Hardware and implementation.}
We use two \(1280\times720\) Intel RealSense D435i cameras for third-person and eye-in-hand views of a 7-DoF Franka arm. All models are implemented in PyTorch with CUDA 11.8 and trained on two NVIDIA A100 GPUs. The Action Policy runs at 15 Hz, with relation checks every five steps.

\begin{figure*}[t] 
  \centering
  \includegraphics[
    page=1,
    trim=0mm 170mm 70mm 0mm,clip,
    width=\textwidth
  ]{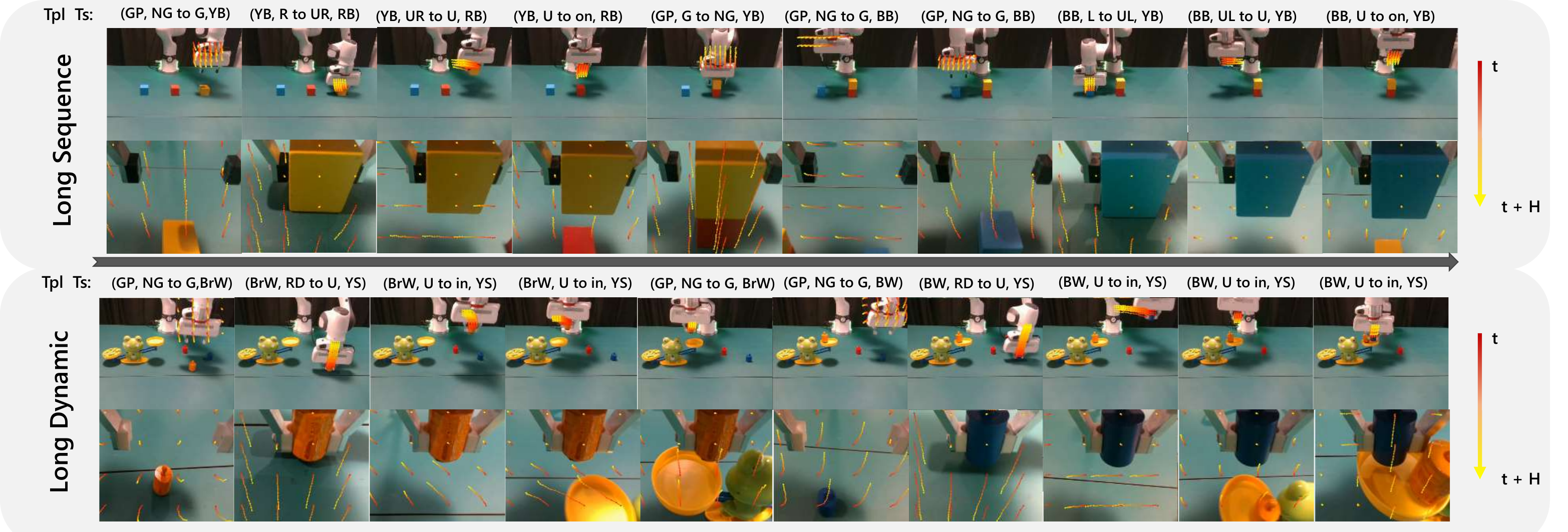} 
  \caption{Partial visualization. Top: triplet transitions from the Task Planner(GP/R/UR/RD/U/UL=gripper/right/upper-right/right-down/upper/upper-left; G/NG=grasp/no-grasp; Y/R/B/Br+B/W/S=yellow/red/blue/Brown+block/weight/scale). Bottom: two-camera point-track views.}
  \label{fig:two_columns}
\end{figure*}

\subsection{Baselines and Results}

We compare TTS with three representative baselines. 
\textbf{FT-Pi0.5}~\cite{intelligence2025pi_} denotes a fine-tuned Pi0.5 model, i.e., an end-to-end VLA baseline that maps images and language directly to actions. We include it to evaluate whether a direct action-generation policy can handle long-horizon tasks with both dexterity and combinatorial structure under our low-data setting. 
\textbf{SeeDo}~\cite{wang2024vlm} is a hierarchical baseline in which a VLM produces a QA-style text plan executed with code-as-policy, representing QA-driven planning from human videos. 
\textbf{PALO}~\cite{myers2024policy} is a semantic planning baseline in which a VLM recursively decomposes a task embedding into executable atomic actions, representing VLM-driven subgoal setting in semantic space. 
For fairness, our Relation Predictor is used only to detect task completion and guard violations, and does not provide additional spatial-relation supervision to the zero-shot VLM planners beyond the initial frame. 

\textbf{Results.}
Across the seven settings in Table I, TTS achieves an average success rate of 74.8\%, compared with 30.4\% for the best-performing baseline, SeeDo.
On \textbf{Diverse pick-and-place}, SeeDo and PALO fail mainly because free-form goal generation is weakly constrained by observations, making them prone to spatial hallucinations such as selecting the wrong object after the interaction state has changed. On \textbf{Sequential stacking}, the dominant failures are logical misordering and drift in multi-step goal tracking, for example grasping an object that is not required at the current stage or executing the prescribed order incorrectly. These errors become more frequent when subgoals are represented only in free-form or weakly grounded semantic space. In contrast, explicit instance-grounded subgoals together with observation-based checking help mitigate these failures.
On \textbf{Dynamic scale balancing}, the main challenge is that the correct next action depends on the evolving physical state after each pick-and-place step. Here, baselines are less stable because they do not explicitly maintain closed-loop planning over changing observations. In the case of FT-Pi0.5, visually similar objects often induce indecision or misselection under our task-specific low-data adaptation setting. Overall, these results indicate that structured, instance-grounded subgoals together with track-based execution priors improve reliability on both seen and unseen long-horizon tasks.

\subsection{Reliability and Stability}

We evaluate reliability and execution stability through controlled perturbation tests, together with qualitative analyses of interpretability and hallucination suppression.

\textbf{Closed-Loop Robustness under Perturbations.}
We test robustness by injecting perturbations during execution. In \textbf{Dynamic scale balancing}, we add or remove frog-shaped weights while keeping the task solvable. In \textbf{Diverse pick-and-place}, we remove the target immediately after grasping to trigger re-localization and re-grasp planning. Success rates under these perturbations are reported in Table~I.

\textbf{Triplets and Tracks for Interpretability and Hallucination Suppression.}
Fig.~5 visualizes representative triplet plans and track trajectories. Because both representations are explicit, execution failures can be traced to the corresponding planning or control stage. We observe that replacing explicit, validatable triplets with a semantic VLM plan (e.g., PALO) leads to a clear drop in performance (Table~I), with failures often caused by planning hallucinations such as incorrect in--out relations, object mislocalization, and redundant picks. Removing the temporal-graph constraint further increases such errors (Sec.~D), suggesting that triplets together with the graph help suppress hallucinated transitions by restricting proposals to observation-consistent and feasible updates.

At execution time, TTS uses object tracks as a physically grounded interface between planning and control. These tracks are explicit and visualizable, keep decision traceable, and support stable closed-loop behavior under limited robot demonstrations. Taken together, these results indicate that triplet-level structure mainly improves planning reliability, while track-based priors help stabilize downstream execution.

\begin{table}[t]
  \centering
  \textbf{TABLE \Rmnum{2}: Planner variants: rollout and overall.}
  \label{tab:planner_variants}
  \vspace{4pt} 
  \footnotesize
  \renewcommand{\arraystretch}{1.05}
  \begin{tabular*}{\linewidth}{@{\extracolsep{\fill}}lcc@{}}
    \toprule
    Method & Rollout Time $\downarrow$ (s) & Overall Accuracy $\uparrow$ (\%) \\
    \midrule
    Only LTP        & $3.4\times 10^{-3}$ & 84.3 \\
    Only VLM        & 202.6               & 98.8 \\
    LTP + VLM       & 32.2                & 98.4 \\
    \bottomrule
  \end{tabular*}
\end{table}

\begin{table}[t]
  \centering
  \textbf{TABLE \Rmnum{3}: Ablation Results: per-family success (\%). }
  \label{tab:ablations_success}
  \vspace{1mm}
  \footnotesize
  \setlength{\tabcolsep}{2pt}
  \renewcommand{\arraystretch}{1.08}

  \begin{tabularx}{\linewidth}{
    @{}
    >{\raggedright\arraybackslash}X
    >{\centering\arraybackslash}p{0.19\linewidth}
    >{\centering\arraybackslash}p{0.17\linewidth}
    >{\centering\arraybackslash}p{0.17\linewidth}
    @{}
  }
    \toprule
    Method/Task & DPP & SS & DSB \\
    \midrule
    w/o Track Predictor
      & 45.5 $\pm$ 7.4 & 46.1 $\pm$ 3.4 & 43.8 $\pm$ 9.4 \\
    w/o Temporal Graph
      & 57.1 $\pm$ 8.4 & 25.0 $\pm$ 0.0 & 0.0 $\pm$ 0.0 \\
    w/o Relation Predictor
      & 75.0 $\pm$ 0.0 & 37.5 $\pm$ 5.5 & 14.3 $\pm$ 0.0 \\
    w/o Task Planner (ATM)
      & 7.1 $\pm$ 2.0 & 0.0 $\pm$ 0.0 & 0.0 $\pm$ 0.0 \\
    Only Action Policy
      & 0.0 $\pm$ 0.0 & 0.0 $\pm$ 0.0 & 0.0 $\pm$ 0.0 \\
    \bottomrule
  \end{tabularx}
\end{table}

\subsection{Design Justification and Ablation Analysis}

We justify the design through real-world ablations that quantify the contribution of each component, including the two relation-generation branches, the temporal graph, task decomposition (Task Planner), trajectory generation (Track Predictor), a policy-only baseline, and the effect of action-labeled data scale on performance and generalization.

\textbf{Effect of Task Planner Components.}
We ablate the relation-generation branches and temporal graph. As shown in Table~\Rmnum{2}, combining both branches improves accuracy with acceptable latency. Removing the graph increases next-triplet errors by making the VLM more prone to hallucinations and dependency-tracking failures, while graph constraints reduce these errors by limiting transitions to feasible updates observed in human data. Table~\Rmnum{3} shows that zero-shot VLMs can infer salient relations in some cases, such as DPP without the Track Predictor, but struggle with fine-grained distinctions required for task switching.

\textbf{Effect of Hierarchical Architecture.}
Layer-wise ablations show that the hierarchy is essential. Without the Task Planner, long-horizon tasks collapse into short-horizon behavior and the system cannot maintain multi-step task progress. Removing the Track Predictor leaves a gap between discrete subgoals and continuous control, which substantially degrades performance. The policy-only baseline (BC) fails consistently because it lacks explicit task-level structure and therefore cannot reliably maintain global progress during execution, as shown in Table~\Rmnum{3}.

\textbf{Effect of Action-Labeled Data Size.}
Finally, we vary the amount of robot action labels used to train the policy and plot the corresponding performance curves. As shown in Fig.~6, TTS achieves reasonable performance with only a few dozen action-labeled demonstrations, supporting its data efficiency.

\begin{figure}  
  \centering
  \includegraphics[
    page=1,
    trim=0mm 295mm 320mm 0mm,clip,
    width=\textwidth
  ]{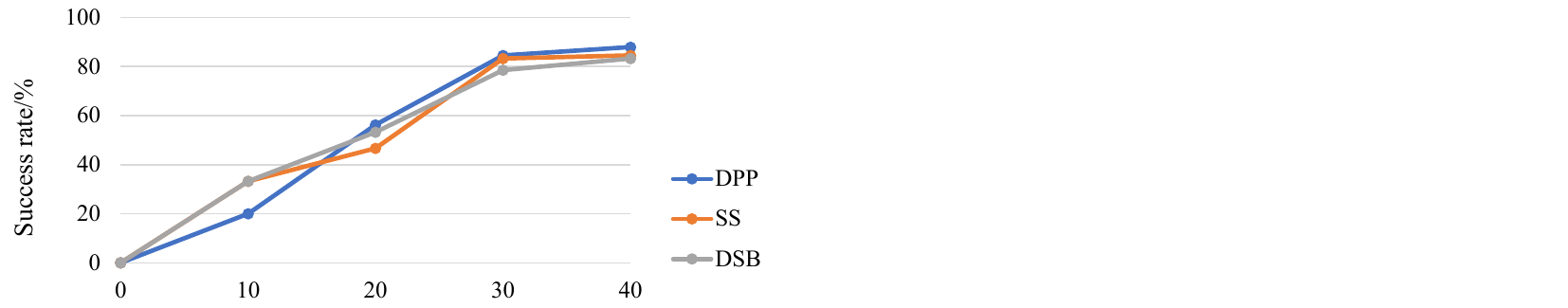} 
  \caption{TTS Success rate with 0–40 robot Demos.}
  \label{fig:two_columns}
\end{figure}




\section{CONCLUSIONS}





We introduced TTS, a closed-loop system using discrete triplets for planning and continuous tracks for execution. Its structured, instance-spatially grounded design reduces hallucinations, improves long-horizon reliability under limited data, and supports object-level and compositional generalization. Limitations include out-of-graph failures, perception errors, and non-triplet tasks such as deformable-object manipulation. Future work will explore state relocalization and richer object representations.











\bibliographystyle{IEEEtran}
\bibliography{myrefs}

@article{wen2024can,
  title={Can Transformers Capture Spatial Relations between Objects?},
  author={Wen, Chuan and Jayaraman, Dinesh and Gao, Yang},
  journal={arXiv preprint arXiv:2403.00729},
  year={2024}
}

@article{chen2024expanding,
  title={Expanding performance boundaries of open-source multimodal models with model, data, and test-time scaling},
  author={Chen, Zhe and Wang, Weiyun and Cao, Yue and Liu, Yangzhou and Gao, Zhangwei and Cui, Erfei and Zhu, Jinguo and Ye, Shenglong and Tian, Hao and Liu, Zhaoyang and others},
  journal={arXiv preprint arXiv:2412.05271},
  year={2024}
}

@article{intelligence2025pi_,
  title={$\pi_{0.5}$: a Vision-Language-Action Model with Open-World Generalization},
  author={Intelligence, Physical and Black, Kevin and Brown, Noah and Darpinian, James and Dhabalia, Karan and Driess, Danny and Esmail, Adnan and Equi, Michael and Finn, Chelsea and Fusai, Niccolo and others},
  journal={arXiv preprint arXiv:2504.16054},
  year={2025}
}

@article{feng2025reflective,
  title={Reflective planning: Vision-language models for multi-stage long-horizon robotic manipulation},
  author={Feng, Yunhai and Han, Jiaming and Yang, Zhuoran and Yue, Xiangyu and Levine, Sergey and Luo, Jianlan},
  journal={arXiv preprint arXiv:2502.16707},
  year={2025}
}

@inproceedings{chen2024spatialvlm,
  title={Spatialvlm: Endowing vision-language models with spatial reasoning capabilities},
  author={Chen, Boyuan and Xu, Zhuo and Kirmani, Sean and Ichter, Brain and Sadigh, Dorsa and Guibas, Leonidas and Xia, Fei},
  booktitle={Proceedings of the IEEE/CVF Conference on Computer Vision and Pattern Recognition},
  pages={14455--14465},
  year={2024}
}

@article{cheng2024spatialrgpt,
  title={Spatialrgpt: Grounded spatial reasoning in vision-language models},
  author={Cheng, An-Chieh and Yin, Hongxu and Fu, Yang and Guo, Qiushan and Yang, Ruihan and Kautz, Jan and Wang, Xiaolong and Liu, Sifei},
  journal={Advances in Neural Information Processing Systems},
  volume={37},
  pages={135062--135093},
  year={2024}
}

@article{black2024pi_0,
  title={$\pi_0$: A Vision-Language-Action Flow Model for General Robot Control},
  author={Black, Kevin and Brown, Noah and Driess, Danny and Esmail, Adnan and Equi, Michael and Finn, Chelsea and Fusai, Niccolo and Groom, Lachy and Hausman, Karol and Ichter, Brian and others},
  journal={arXiv preprint arXiv:2410.24164},
  year={2024}
}

@article{kim2024openvla,
  title={Openvla: An open-source vision-language-action model},
  author={Kim, Moo Jin and Pertsch, Karl and Karamcheti, Siddharth and Xiao, Ted and Balakrishna, Ashwin and Nair, Suraj and Rafailov, Rafael and Foster, Ethan and Lam, Grace and Sanketi, Pannag and others},
  journal={arXiv preprint arXiv:2406.09246},
  year={2024}
}

@article{wang2024vlm,
  title={Vlm see, robot do: Human demo video to robot action plan via vision language model},
  author={Wang, Beichen and Zhang, Juexiao and Dong, Shuwen and Fang, Irving and Feng, Chen},
  journal={arXiv preprint arXiv:2410.08792},
  year={2024}
}

@article{zhang2024hirt,
  title={Hirt: Enhancing robotic control with hierarchical robot transformers},
  author={Zhang, Jianke and Guo, Yanjiang and Chen, Xiaoyu and Wang, Yen-Jen and Hu, Yucheng and Shi, Chengming and Chen, Jianyu},
  journal={arXiv preprint arXiv:2410.05273},
  year={2024}
}

@article{myers2024policy,
  title={Policy adaptation via language optimization: Decomposing tasks for few-shot imitation},
  author={Myers, Vivek and Zheng, Bill Chunyuan and Mees, Oier and Levine, Sergey and Fang, Kuan},
  journal={arXiv preprint arXiv:2408.16228},
  year={2024}
}

@article{li2025gr,
  title={Gr-mg: Leveraging partially-annotated data via multi-modal goal-conditioned policy},
  author={Li, Peiyan and Wu, Hongtao and Huang, Yan and Cheang, Chilam and Wang, Liang and Kong, Tao},
  journal={IEEE Robotics and Automation Letters},
  year={2025},
  publisher={IEEE}
}

@article{liu2025aligning,
  title={Aligning cyber space with physical world: A comprehensive survey on embodied ai},
  author={Liu, Yang and Chen, Weixing and Bai, Yongjie and Liang, Xiaodan and Li, Guanbin and Gao, Wen and Lin, Liang},
  journal={IEEE/ASME Transactions on Mechatronics},
  year={2025},
  publisher={IEEE}
}

@article{duan2022survey,
  title={A survey of embodied ai: From simulators to research tasks},
  author={Duan, Jiafei and Yu, Samson and Tan, Hui Li and Zhu, Hongyuan and Tan, Cheston},
  journal={IEEE Transactions on Emerging Topics in Computational Intelligence},
  volume={6},
  number={2},
  pages={230--244},
  year={2022},
  publisher={IEEE}
}

@article{wang2024karma,
  title={Karma: Augmenting embodied ai agents with long-and-short term memory systems},
  author={Wang, Zixuan and Yu, Bo and Zhao, Junzhe and Sun, Wenhao and Hou, Sai and Liang, Shuai and Hu, Xing and Han, Yinhe and Gan, Yiming},
  journal={arXiv preprint arXiv:2409.14908},
  year={2024}
}

@inproceedings{yu2020meta,
  title={Meta-world: A benchmark and evaluation for multi-task and meta reinforcement learning},
  author={Yu, Tianhe and Quillen, Deirdre and He, Zhanpeng and Julian, Ryan and Hausman, Karol and Finn, Chelsea and Levine, Sergey},
  booktitle={Conference on robot learning},
  pages={1094--1100},
  year={2020},
  organization={PMLR}
}

@article{james2020rlbench,
  title={Rlbench: The robot learning benchmark \& learning environment},
  author={James, Stephen and Ma, Zicong and Arrojo, David Rovick and Davison, Andrew J},
  journal={IEEE Robotics and Automation Letters},
  volume={5},
  number={2},
  pages={3019--3026},
  year={2020},
  publisher={IEEE}
}

@inproceedings{zeng2021transporter,
  title={Transporter networks: Rearranging the visual world for robotic manipulation},
  author={Zeng, Andy and Florence, Pete and Tompson, Jonathan and Welker, Stefan and Chien, Jonathan and Attarian, Maria and Armstrong, Travis and Krasin, Ivan and Duong, Dan and Sindhwani, Vikas and others},
  booktitle={Conference on Robot Learning},
  pages={726--747},
  year={2021},
  organization={PMLR}
}

@misc{yang2025lohovlaunifiedvisionlanguageactionmodel,
      title={LoHoVLA: A Unified Vision-Language-Action Model for Long-Horizon Embodied Tasks}, 
      author={Yi Yang and Jiaxuan Sun and Siqi Kou and Yihan Wang and Zhijie Deng},
      year={2025},
      eprint={2506.00411},
      archivePrefix={arXiv},
}

@article{zhou2025step,
  title={STEP Planner: Constructing cross-hierarchical subgoal tree as an embodied long-horizon task planner},
  author={Zhou, Tianxing and Wang, Zhirui and Ao, Haojia and Chen, Guangyan and Xing, Boyang and Cheng, Jingwen and Yang, Yi and Yue, Yufeng},
  journal={arXiv preprint arXiv:2506.21030},
  year={2025}
}

@article{cheang2024gr,
  title={Gr-2: A generative video-language-action model with web-scale knowledge for robot manipulation},
  author={Cheang, Chi-Lam and Chen, Guangzeng and Jing, Ya and Kong, Tao and Li, Hang and Li, Yifeng and Liu, Yuxiao and Wu, Hongtao and Xu, Jiafeng and Yang, Yichu and others},
  journal={arXiv preprint arXiv:2410.06158},
  year={2024}
}

@article{wen2023any,
  title={Any-point trajectory modeling for policy learning},
  author={Wen, Chuan and Lin, Xingyu and So, John and Chen, Kai and Dou, Qi and Gao, Yang and Abbeel, Pieter},
  journal={arXiv preprint arXiv:2401.00025},
  year={2023}
}

@article{brohan2022rt,
  title={Rt-1: Robotics transformer for real-world control at scale},
  author={Brohan, Anthony and Brown, Noah and Carbajal, Justice and Chebotar, Yevgen and Dabis, Joseph and Finn, Chelsea and Gopalakrishnan, Keerthana and Hausman, Karol and Herzog, Alex and Hsu, Jasmine and others},
  journal={arXiv preprint arXiv:2212.06817},
  year={2022}
}

@article{gao2025vla,
  title={VLA-OS: Structuring and Dissecting Planning Representations and Paradigms in Vision-Language-Action Models},
  author={Gao, Chongkai and Liu, Zixuan and Chi, Zhenghao and Huang, Junshan and Fei, Xin and Hou, Yiwen and Zhang, Yuxuan and Lin, Yudi and Fang, Zhirui and Jiang, Zeyu and others},
  journal={arXiv preprint arXiv:2506.17561},
  year={2025}
}

@article{wang2025vlatest,
  title={VLATest: Testing and Evaluating Vision-Language-Action Models for Robotic Manipulation},
  author={Wang, Zhijie and Zhou, Zhehua and Song, Jiayang and Huang, Yuheng and Shu, Zhan and Ma, Lei},
  journal={Proceedings of the ACM on Software Engineering},
  volume={2},
  number={FSE},
  pages={1615--1638},
  year={2025},
  publisher={ACM New York, NY, USA}
}

@article{ahn2022can,
  title={Do as i can, not as i say: Grounding language in robotic affordances},
  author={Ahn, Michael and Brohan, Anthony and Brown, Noah and Chebotar, Yevgen and Cortes, Omar and David, Byron and Finn, Chelsea and Fu, Chuyuan and Gopalakrishnan, Keerthana and Hausman, Karol and others},
  journal={arXiv preprint arXiv:2204.01691},
  year={2022}
}

@inproceedings{wu2024mldt,
  title={Mldt: Multi-level decomposition for complex long-horizon robotic task planning with open-source large language model},
  author={Wu, Yike and Zhang, Jiatao and Hu, Nan and Tang, Lanling and Qi, Guilin and Shao, Jun and Ren, Jie and Song, Wei},
  booktitle={International Conference on Database Systems for Advanced Applications},
  pages={251--267},
  year={2024},
  organization={Springer}
}

@article{xu2024flow,
  title={Flow as the cross-domain manipulation interface},
  author={Xu, Mengda and Xu, Zhenjia and Xu, Yinghao and Chi, Cheng and Wetzstein, Gordon and Veloso, Manuela and Song, Shuran},
  journal={arXiv preprint arXiv:2407.15208},
  year={2024}
}

@article{bharadhwaj2024track2act,
  title={Track2act: Predicting point tracks from internet videos enables diverse zero-shot robot manipulation},
  author={Bharadhwaj, Homanga and Mottaghi, Roozbeh and Gupta, Abhinav and Tulsiani, Shubham},
  journal={CoRR},
  year={2024}
}

@article{zhang2024vlabench,
  title={Vlabench: A large-scale benchmark for language-conditioned robotics manipulation with long-horizon reasoning tasks},
  author={Zhang, Shiduo and Xu, Zhe and Liu, Peiju and Yu, Xiaopeng and Li, Yuan and Gao, Qinghui and Fei, Zhaoye and Yin, Zhangyue and Wu, Zuxuan and Jiang, Yu-Gang and others},
  journal={arXiv preprint arXiv:2412.18194},
  year={2024}
}

@article{fan2025long,
  title={Long-VLA: Unleashing Long-Horizon Capability of Vision Language Action Model for Robot Manipulation},
  author={Fan, Yiguo and Ding, Pengxiang and Bai, Shuanghao and Tong, Xinyang and Zhu, Yuyang and Lu, Hongchao and Dai, Fengqi and Zhao, Wei and Liu, Yang and Huang, Siteng and others},
  journal={arXiv preprint arXiv:2508.19958},
  year={2025}
}

@article{sun2024hierarchical,
  title={Hierarchical hybrid learning for long-horizon contact-rich robotic assembly},
  author={Sun, Jiankai and Curtis, Aidan and You, Yang and Xu, Yan and Koehle, Michael and Guibas, Leonidas and Chitta, Sachin and Schwager, Mac and Li, Hui},
  journal={arXiv preprint arXiv:2409.16451},
  year={2024}
}

@article{sundaresan2024s,
  title={What's the Move? Hybrid Imitation Learning via Salient Points},
  author={Sundaresan, Priya and Hu, Hengyuan and Vuong, Quan and Bohg, Jeannette and Sadigh, Dorsa},
  journal={arXiv preprint arXiv:2412.05426},
  year={2024}
}

@inproceedings{zhou2024isr,
  title={Isr-llm: Iterative self-refined large language model for long-horizon sequential task planning},
  author={Zhou, Zhehua and Song, Jiayang and Yao, Kunpeng and Shu, Zhan and Ma, Lei},
  booktitle={2024 IEEE International Conference on Robotics and Automation (ICRA)},
  pages={2081--2088},
  year={2024},
  organization={IEEE}
}

@article{shao2021concept2robot,
  title={Concept2robot: Learning manipulation concepts from instructions and human demonstrations},
  author={Shao, Lin and Migimatsu, Toki and Zhang, Qiang and Yang, Karen and Bohg, Jeannette},
  journal={The International Journal of Robotics Research},
  volume={40},
  number={12-14},
  pages={1419--1434},
  year={2021},
  publisher={SAGE Publications Sage UK: London, England}
}

@inproceedings{shaw2023videodex,
  title={Videodex: Learning dexterity from internet videos},
  author={Shaw, Kenneth and Bahl, Shikhar and Pathak, Deepak},
  booktitle={Conference on Robot Learning},
  pages={654--665},
  year={2023},
  organization={PMLR}
}

@article{baker2022video,
  title={Video pretraining (vpt): Learning to act by watching unlabeled online videos},
  author={Baker, Bowen and Akkaya, Ilge and Zhokov, Peter and Huizinga, Joost and Tang, Jie and Ecoffet, Adrien and Houghton, Brandon and Sampedro, Raul and Clune, Jeff},
  journal={Advances in Neural Information Processing Systems},
  volume={35},
  pages={24639--24654},
  year={2022}
}

@article{torabi2018behavioral,
  title={Behavioral cloning from observation},
  author={Torabi, Faraz and Warnell, Garrett and Stone, Peter},
  journal={arXiv preprint arXiv:1805.01954},
  year={2018}
}

@article{nair2022r3m,
  title={R3m: A universal visual representation for robot manipulation},
  author={Nair, Suraj and Rajeswaran, Aravind and Kumar, Vikash and Finn, Chelsea and Gupta, Abhinav},
  journal={arXiv preprint arXiv:2203.12601},
  year={2022}
}

@article{ma2022vip,
  title={Vip: Towards universal visual reward and representation via value-implicit pre-training},
  author={Ma, Yecheng Jason and Sodhani, Shagun and Jayaraman, Dinesh and Bastani, Osbert and Kumar, Vikash and Zhang, Amy},
  journal={arXiv preprint arXiv:2210.00030},
  year={2022}
}

@inproceedings{sermanet2018time,
  title={Time-contrastive networks: Self-supervised learning from video},
  author={Sermanet, Pierre and Lynch, Corey and Chebotar, Yevgen and Hsu, Jasmine and Jang, Eric and Schaal, Stefan and Levine, Sergey and Brain, Google},
  booktitle={2018 IEEE international conference on robotics and automation (ICRA)},
  pages={1134--1141},
  year={2018},
  organization={IEEE}
}

@article{ko2023learning,
  title={Learning to act from actionless videos through dense correspondences},
  author={Ko, Po-Chen and Mao, Jiayuan and Du, Yilun and Sun, Shao-Hua and Tenenbaum, Joshua B},
  journal={arXiv preprint arXiv:2310.08576},
  year={2023}
}

@article{bharadhwaj2023zero,
  title={Zero-shot robot manipulation from passive human videos},
  author={Bharadhwaj, Homanga and Gupta, Abhinav and Tulsiani, Shubham and Kumar, Vikash},
  journal={arXiv preprint arXiv:2302.02011},
  year={2023}
}

@inproceedings{liu2024grounding,
  title={Grounding dino: Marrying dino with grounded pre-training for open-set object detection},
  author={Liu, Shilong and Zeng, Zhaoyang and Ren, Tianhe and Li, Feng and Zhang, Hao and Yang, Jie and Jiang, Qing and Li, Chunyuan and Yang, Jianwei and Su, Hang and others},
  booktitle={European conference on computer vision},
  pages={38--55},
  year={2024},
  organization={Springer}
}

@article{ravi2024sam,
  title={Sam 2: Segment anything in images and videos},
  author={Ravi, Nikhila and Gabeur, Valentin and Hu, Yuan-Ting and Hu, Ronghang and Ryali, Chaitanya and Ma, Tengyu and Khedr, Haitham and R{\"a}dle, Roman and Rolland, Chloe and Gustafson, Laura and others},
  journal={arXiv preprint arXiv:2408.00714},
  year={2024}
}

@inproceedings{islam2025gpt,
  title={Gpt-4o: The cutting-edge advancement in multimodal llm},
  author={Islam, Raisa and Moushi, Owana Marzia},
  booktitle={Intelligent Computing-Proceedings of the Computing Conference},
  pages={47--60},
  year={2025},
  organization={Springer}
}

@article{karaev2024cotracker3,
  title={Cotracker3: Simpler and better point tracking by pseudo-labelling real videos},
  author={Karaev, Nikita and Makarov, Iurii and Wang, Jianyuan and Neverova, Natalia and Vedaldi, Andrea and Rupprecht, Christian},
  journal={arXiv preprint arXiv:2410.11831},
  year={2024}
}

\end{document}